\documentclass[runningheads]{llncs}

\usepackage[mobile]{eccv}

\usepackage{eccvabbrv}

\usepackage{multirow}
\usepackage{graphicx}
\usepackage{booktabs}
\usepackage{array}
\usepackage{pifont}
\usepackage{xcolor}
\usepackage{color, colortbl}	
\definecolor{Lavender}{rgb}{0.8,0.84,0.98}
\definecolor{Cream}{rgb}{0.96,0.96,0.96}
\definecolor{DarkCream}{rgb}{0.9,0.9,0.9}
\newcommand{\cmark}{\textcolor{green!80!black}{\ding{51}}}
\newcommand{\xmark}{\textcolor{red}{\ding{55}}}
\usepackage{arydshln}

\usepackage[accsupp]{axessibility}  

\usepackage[pagebackref,breaklinks,colorlinks]{hyperref}
\usepackage{hyperref}

\usepackage{orcidlink}

\begin{document}\renewcommand{\arraystretch}{1.2}

\title{PARIS3D: Reasoning-based 3D Part Segmentation Using Large Multimodal Model} 

\titlerunning{PARIS3D: Reasoning-based 3D Part Segmentation}

\author{Amrin Kareem\inst{1} \and
Jean Lahoud\inst{1} \and
Hisham Cholakkal \inst{1}}

\authorrunning{~Kareem et al.}

\institute{Mohamed bin Zayed University of Artificial Intelligence, Abu Dhabi, UAE} 

\maketitle

\begin{abstract}
  Recent advancements in 3D perception systems have significantly improved their ability to perform visual recognition tasks such as segmentation. However, these systems still heavily rely on explicit human instruction to identify target objects or categories, lacking the capability to actively reason and comprehend implicit user intentions. We introduce a novel segmentation task known as reasoning part segmentation for 3D objects, aiming to output a segmentation mask based on complex and implicit textual queries about specific parts of a 3D object.  To facilitate evaluation and benchmarking, we present a large 3D dataset comprising over 60k instructions paired with corresponding ground-truth part segmentation annotations specifically curated for reasoning-based 3D part segmentation.  We propose a model that is capable of segmenting parts of 3D objects based on implicit textual queries and generating natural language explanations corresponding to 3D object segmentation requests. Experiments show that our method achieves competitive performance to models that use explicit queries, with the additional abilities to identify part concepts, reason about them, and complement them with world knowledge. Our source code, dataset, and trained models are available \href{https://github.com/AmrinKareem/PARIS3D}{here}. 
\end{abstract}

\section{Introduction}
\label{sec:intro}

The rapid advancements in 3D data capture technologies, including LIDARs and RGB-D cameras, have led to a growing demand for automated analysis of 3D point clouds. 3D semantic segmentation, the process of automatically assigning predefined semantic labels to each point in a cloud, is crucial for enabling complex tasks such as scene understanding. Similarly, 3D part segmentation involves further segmenting object instances into their components, such as identifying the handle of a pot or the lid of a bottle. These tasks find applications in various fields including autonomous vehicles, mobile robotics, industrial automation, augmented reality, and medical imagery analysis.
\begin{figure}[t]
    \centering
    \includegraphics[width=1\linewidth]{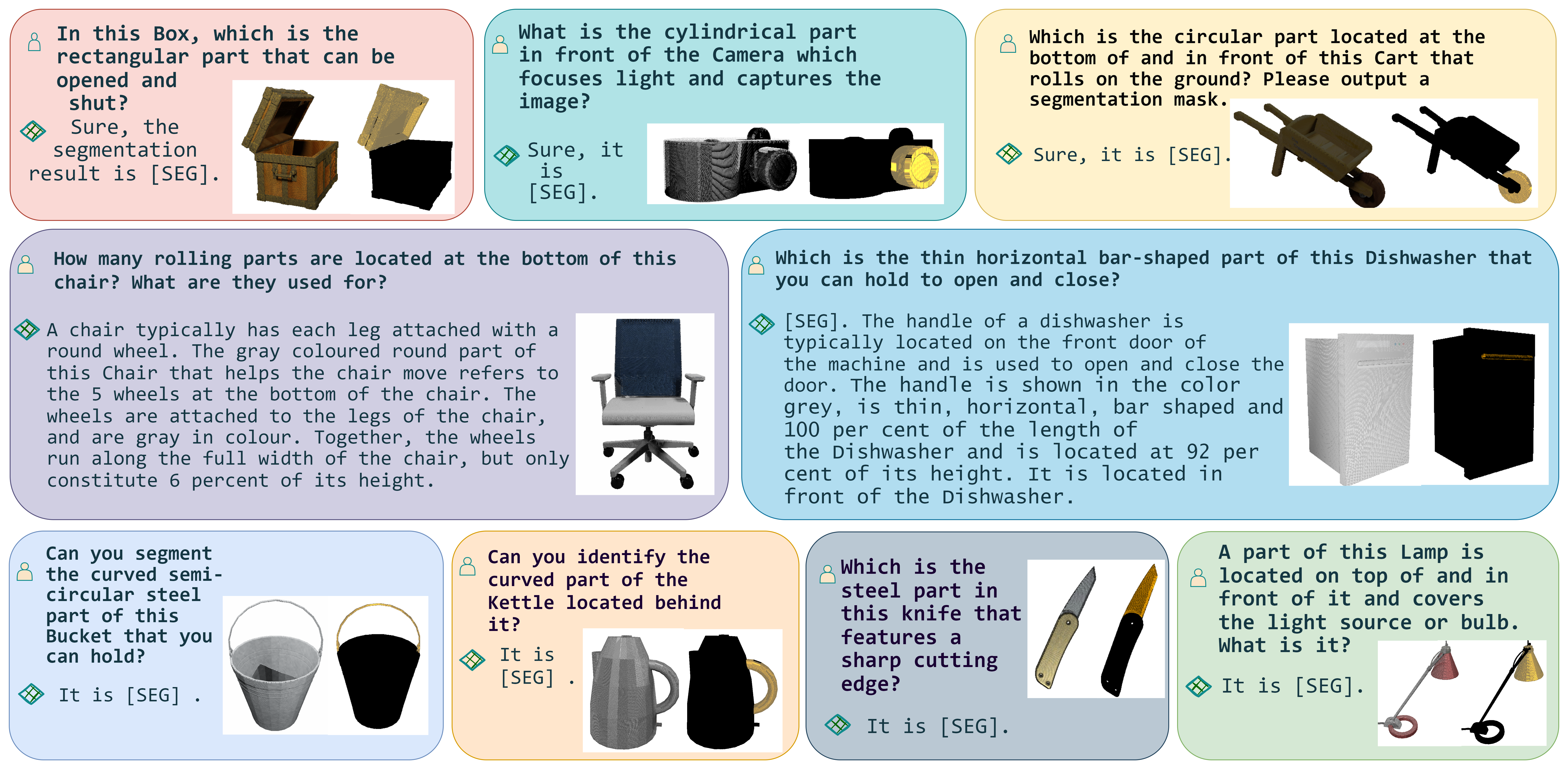}
    \caption{\textbf{Capabilities of PARIS3D.} Parts of 3D objects are segmented based on reasoning, shape, location, material, colour, and concept instructions. Additionally, for the segmentations, PARIS3D can explain why it chose that region, or describe 3D objects with respect to their parts. The original point clouds are on the left. The segmented parts are shown to the right, highlighted in golden colour.}
    \label{fig:intro}
\end{figure}
While recent advancements in pre-trained 3D representations and the introduction of various 3D datasets have significantly improved 3D perception, the capacity for nuanced reasoning in 3D contexts remains limited. This limitation is primarily due to the lack of comprehensive datasets for reasoning and describing 3D scenes and objects. In contrast, 2D images accompanied by textual descriptions have contributed to significant progress in large-scale image language models. Recent methods have emerged, performing segmentation based on 2D images projected from 3D scenes. However, they lack the ability to reason about object concepts and their 3D properties, functions, or parts in a conversational manner.

Here, we define a model as having reasoning capabilities when (i) it can understand implicit instructions, such as referencing properties of an object or its parts without explicit articulation, and (ii) explain or justify its responses, whether generated text or predicted segmentation masks. Such reasoning ability is one of the fundamental cognitive skills possessed by humans, essential for daily activities ranging from locating items to manipulating tools. However, prevailing perception systems often require explicit human input to designate target objects or predefined categories before performing visual recognition tasks. These systems cannot autonomously deduce and comprehend users' intentions based on implicit instructions. The integration of self-reasoning capabilities is key to advancing the next generation of intelligent perception systems, with considerable potential for diverse  applications including robotics.

In this work, we introduce a novel task termed reasoning part segmentation in 3D. This task involves generating a part segmentation mask for a 3D object based on implicit textual queries requiring complex reasoning. These queries go beyond simple references and encompass intricate descriptions, demanding sophisticated reasoning or worldly knowledge. For example, knowing where to hold a kettle or which part of a bottle to open in a fine-grained manner. 
To address this challenging task, we propose PARIS3D (\textbf{R}eason\textbf{I}ng-based \textbf{3D} \textbf{PA}rt \textbf{S}egmentation), a multimodal Large Language Model (LLM) capable of reasoning on user input text, predicting 3D part segmentation masks, and providing explanations for the model's response. As depicted in Figure \ref{fig:intro}, PARIS3D adeptly navigates diverse scenarios, encompassing complex 3D reasoning, material, color, shape, and location-based knowledge, providing explanatory responses and detailed descriptions. Additionally, to validate the effectiveness of PARIS3D and support future research, we establish an evaluation benchmark and a dataset named RPSeg3D for reasoning 3D part segmentation. Our RPSeg3D dataset comprises 2624 3D objects and over 60k instructions, providing persuasive evaluation metrics for the task.

In summary, our key contributions are the following:
\begin{itemize}
\item We introduce the reasoning part segmentation task for 3D objects, emphasizing the necessity of reasoning capabilities which is crucial for the development of intelligent perception systems. 
    \item We provide a comprehensive dataset named RPSeg3D for reasoning-based 3D part segmentation, comprising over 2624 3D objects and 60k instructions, serving as a useful resource for future research.  
    \item We present PARIS3D, a novel approach for 3D part segmentation, and further improve its capabilities by fine-tuning on our RPSeg3D dataset.
\end{itemize}

\section{Related Work} 
\noindent\textbf{3D Semantic Segmentation.}
The challenge of understanding and reasoning within 3D environments has been an ongoing research focus. The goal of 3D semantic segmentation is to acquire semantic predictions for each point in a cloud. Notable contributions include point-based approaches \cite{qi2017pointnet,8578376}, methods which incorporate intricately crafted point convolution methods \cite{thomas2019KPConv,xu2021paconv}, voxel-based strategies \cite{8579059,8953494}, including those that employ 3D sparse convolutions \cite{graham2017submanifold} for generating point-wise segmentation outcomes, as well as transformer-based techniques \cite{lai2022stratified}. 
Multi-view semantic segmentation methods such as DeepViewAgg \cite{robert2022dva}, Diffuser \cite{9561801,kundu2020virtual}, 3D-CG \cite{hong20223dcg}, 3D-CLR \cite{hong2023threedclr} in 3D vision concentrate on improving representation learning by generating 2D renderings from 3D under multiple view points. These works have shown the effectiveness of multi-view representations in enhancing the performance and robustness of various 3D tasks.
Nonetheless, these methods rely on a predefined set of semantic labels, whereas we focus in our proposed method on responding to complex reasoning-based queries and explaining them. 


\noindent\textbf{Large Multimodal Models.}
Extensive research on Large Language Models (LLMs) has demonstrated reasoning capabilities, prompting an exploration into extending these skills into the visual domain through Large Multimodal Models.
LMMs are highly adaptable and versatile means to perform tasks requiring language and vision capabilities. Prominent models such as BLIP-2 \cite{li2023blip2}, LLaVA \cite{liu2023llava} and MiniGPT-4 \cite{zhu2023minigpt4} generally employ a dual-phase training process,  aligning visual representations with the linguistic embeddings of LLMs through extensive image-text and video-text datasets \cite{lin2015microsoft,sharma2018conceptual,changpinyo2021conceptual,schuhmann2021laion400m,schuhmann2022laion5b,bain2022frozen,miech2019howto100m}. Recent efforts have focused on the convergence of multimodal LLMs with vision tasks, where VisionLLM \cite{wang2023visionllm} provides a versatile interaction interface for a spectrum of vision-centric tasks through instruction tuning. Yet, it does not fully leverage the complex reasoning potential of LLMs. Kosmos-2 \cite{peng2023kosmos2} aims to enrich LLMs with grounding capabilities by creating a large dataset of grounded image-text pairs. DetGPT \cite{pi2023detgpt} seamlessly connects a fixed multimodal LLM framework with an open-vocabulary detector to facilitate instruction-based detection. LISA \cite{lai2023lisa} uses embeddings from the vision language model and the SAM \cite{kirillov2023segment} decoder to generate segmentation masks. GPT4RoI \cite{zhang2023gpt4roi} integrates spatial boxes as inputs and trains models on region-text pairs, showcasing a novel approach. Our method aims to benefit from these advances in the LMM space by merging the vision-language abilities of LMMs and the reasoning of LLMs in a novel 3D perception task.

\noindent\textbf{Language Instructed 3D Tasks.}
The integration of point clouds with natural language processing has widespread implications, drawing considerable interest in the realms of 3D scene understanding. This fast-growing field promises enhancements in human-robot interaction, metaverse, robotics, and embodied intelligence. 
Central to the dialogue systems designed for 3D environments are two critical capabilities: perception within three-dimensional spaces and reasoning. 
\begin{table}[t]
\centering
\scriptsize
\begin{tabular}{c|cc|c|c|c}
\hline
\multirow{2}{*}{Method} & \multicolumn{2}{c}{Input}     & \multicolumn{3}{|c}{Task} \\ \cline{2-3} \cline{4-6}
 &
  \multicolumn{1}{c|}{Point} &
  Reasoning  &
  \multicolumn{1}{c|}{\multirow{2}{*}{Conversation}} &
  \multicolumn{1}{c|}{\multirow{2}{*}{Segmentation}} &
  \multirow{2}{*}{Explanation}
   \\ 
& \multicolumn{1}{c|}{Cloud}   & Query & & &  \\ \hline

  \rowcolor{Cream}
SQA3D    \cite{ma2023sqa3d}, 3D-VisTA   \cite{3dvista}              & \multicolumn{1}{c|}{\cmark} & \cmark & \multicolumn{1}{c|}{\xmark}  & \multicolumn{1}{c|}{\xmark}  & \xmark                              \\ 
 
ViewRefer     \cite{Guo_2023_ICCV}, Point-Bind   \cite{guo2023pointbind}         & \multicolumn{1}{c|}{\cmark} & \xmark  & \multicolumn{1}{c|}{\xmark}  & \multicolumn{1}{c|}{\xmark}  & \xmark                               \\ \hdashline
  \rowcolor{Cream}
3D-OVS    \cite{NEURIPS2023_a76b693f}             & \multicolumn{1}{c|}{\xmark}  & \xmark  & \multicolumn{1}{c|}{\xmark}  & \multicolumn{1}{c|}{\cmark} & \xmark                              \\
OpenMask3D  \cite{takmaz2023openmask3d}           & \multicolumn{1}{c|}{\cmark} & \xmark  & \multicolumn{1}{c|}{\xmark}  & \multicolumn{1}{c|}{\cmark} & \xmark                              \\ 
  \rowcolor{Cream}
PLA    \cite{ding2022language},  OpenScene \cite{Peng2023OpenScene}              & \multicolumn{1}{c|}{\cmark} & \xmark  & \multicolumn{1}{c|}{\xmark}  & \multicolumn{1}{c|}{\cmark} & \xmark                             \\ \hdashline

Chat-3D  \cite{wang2023chat}              & \multicolumn{1}{c|}{\cmark} & \cmark & \multicolumn{1}{c|}{\cmark} & \multicolumn{1}{c|}{\xmark}  & \cmark                             \\ 
  \rowcolor{Cream}
M3DBench  \cite{li2023m3dbench}             & \multicolumn{1}{c|}{\cmark} & \xmark  & \multicolumn{1}{c|}{\cmark}    & \multicolumn{1}{c|}{\xmark}  & \xmark                           \\
LLM-Grounder  \cite{yang2023llmgrounder}         & \multicolumn{1}{c|}{\cmark} & \xmark  & \multicolumn{1}{c|}{\cmark} & \multicolumn{1}{c|}{\xmark}  & \cmark                              \\ 
            \rowcolor{Cream}
3D-LLM    \cite{3dllm}             & \multicolumn{1}{c|}{\cmark} & \cmark & \multicolumn{1}{c|}{\cmark} & \multicolumn{1}{c|}{\xmark}  & \cmark                              \\ 
LL3DA   \cite{chen2023ll3da}, PointLLM \cite{xu2023pointllm}               & \multicolumn{1}{c|}{\cmark} & \xmark  & \multicolumn{1}{c|}{\cmark}    & \multicolumn{1}{c|}{\xmark}  & \cmark                         \\ 

\rowcolor{Lavender}
PARIS3D                & \multicolumn{1}{c|}{\cmark} & \cmark & \multicolumn{1}{c|}{\cmark} & \multicolumn{1}{c|}{\cmark} & \cmark \\
\hline
\end{tabular}%
\caption{\textbf{Comparison of recent 3D segmentation models and Large Multimodal Models (LMMs)} emphasizing their capabilities for 3D reasoning and conversations. Reasoning query means the model is asked to self-reason a task and either output text or perform an action. Segmentation highlights models that can respond with 3D segmentation masks, and Conversation represents models that can provide a conversation-style answer to the user. Among these, our proposed PARIS3D stands out with comprehensive 3D understanding and reasoning, segmentation in response to natural language queries, and conversational capabilities.}
\label{tab:Literature}
\end{table}

Recently, there has been a rise in the number of tasks uniting 3D scenes and language, such as 3D captioning, 3D question answering, 3D situated question answering, embodied Q and A, planning, navigation, 3D assisted multi-turn dialogue, 3D object detection, and scene description. We divide the 3D perception task models into 3 categories (see Table \ref{tab:Literature}. Separated by dotted lines.). The first encompasses models that perform tasks like 3D captioning, situated question answering, and visual grounding\cite{3dvista,ma2023sqa3d}, visual grounding \cite{guo2023viewrefer}. These models are capable of providing a word or phrase as its text output. The second category has 3D semantic segmentation models that output 3D segmentation masks. 3D-OVS \cite{NEURIPS2023_a76b693f}, Openmask3D \cite{takmaz2023openmask3d}, OpenScene \cite{Peng2023OpenScene}, PLA \cite{Ding_2023_CVPR}, perform open-vocabulary semantic segmentation for 3D scenes. These methods, on the other hand, cannot provide a conversational output to a user query or explain the reasoning for their tasks.
The third comprises models that employ an LLM and perform visual perception tasks such as captioning, scene understanding, and visual grounding, providing conversational outputs \cite{xu2023pointllm,guo2023pointbind,3dllm,wang2023chat,li2023m3dbench,yang2023llmgrounder,chen2023ll3da}. However, they do not perform fine-grained semantic segmentation or reasoning-based 3D vision tasks.

As such, we identify a gap in performing 3D segmentation in response to complex natural language prompts instead of a single phrase or word. 
Distinct from the existing works in this domain, our research is committed to:
(1) streamlining the integration of 3D segmentation capabilities into multimodal LLMs, and
(2) enhancing current perception systems with 3D reasoning-based segmentation and explanation abilities, thereby pushing the boundaries of what is achievable in the intersection of language and visual understanding.

\section{Reasoning 3D Part Segmentation}
\label{sec:method}
Semantic segmentation involves assigning a semantic label to each geometric primitive, such as points \cite{qi2017pointnet}, voxels \cite{graham20173d}, or superpoints \cite{landrieu2018largescale}. In part segmentation, object instances are decomposed into their components.
Given a coloured point cloud P, the goal of a 3D segmentation model is to predict its label for each point. However, in our reasoning segmentation task, we go further to output a 3D segmentation mask M, given an input point cloud and an implicit query text instruction $x_{txt}$. 
The task shares a similar formulation with the referring segmentation task \cite{kazemzadeh-etal-2014-referitgame}, with an additional challenge for the model to reason about the fine-grained parts in response to implicit queries and output the corresponding segmentation mask.
The complexity of the query text in reasoning part segmentation is a key differentiator. Instead of providing the names of the parts, the query text may include more intricate expressions that involve an understanding of structures, geometries, and semantics of 3D objects. By introducing this task, we aim to bridge the gap between user intent and system response, enabling more intuitive and dynamic interactions in 3D object perception.  

\begin{figure}[t]
    \centering
\includegraphics[width=1\linewidth]{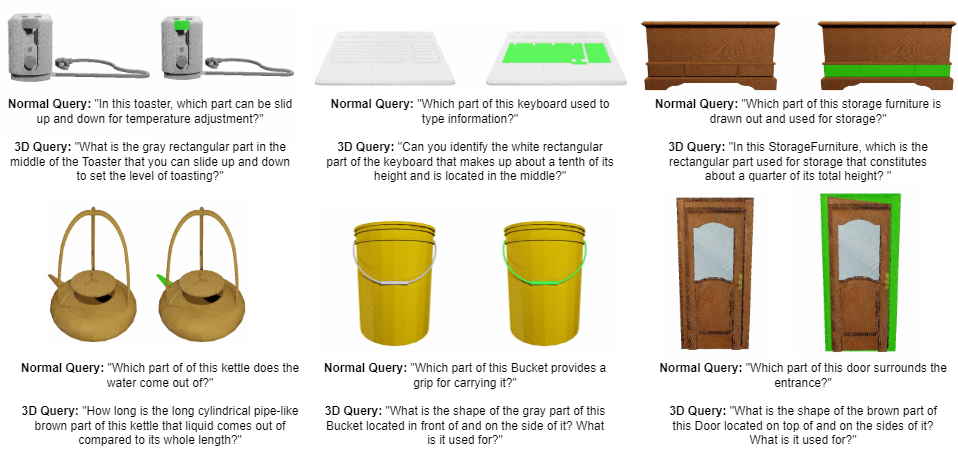}
     \caption{Examples of the annotated object-instruction pairs for training with two types of queries. On the left is one view of the rendered image from the original point cloud. On the right is the corresponding ground truth segmentation mask, shown in green.}
     \label{fig:examplesquery}
\end{figure}
\subsection{Our RPSeg3D Dataset }
\label{subsec:dataset}
\begin{figure}[t]
    \centering
\includegraphics[width=1\linewidth]{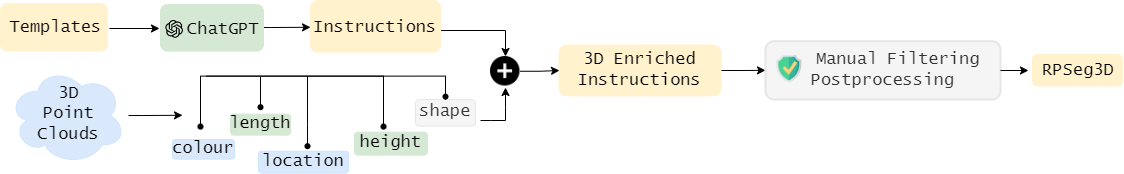}
    \caption{\textbf{Preparing the instructions of RPSeg3D.} Simple templates are provided to GPT-3.5, which populates them with part information-related segmentation instructions. In parallel, colour, shape, location, and dimension-related data is extracted from 3D point clouds. Enriching the instructions with this information and manually checking them for inaccuracies, we obtain the RPSeg3D dataset for part segmentation. }
    \label{fig:dataanno}
\end{figure}

Considering the unavailability of established datasets and evaluation benchmarks in the literature,  we introduce a dataset, named  RPSeg3D,  specifically designed for the reasoning 3D part segmentation task. Our dataset comprises 2624 3D objects and over 60k instructions. We use 718 objects and their corresponding instructions as the train set, and the remaining 1906 objects along with their instructions are used for testing. For reliable and fair assessment, we have aligned the 3D objects with those from PartNet-Ensemble \cite{Liu_2023_CVPR}, annotating them with implicit text instructions and using ground truth labels to generate high-quality target masks. To generate the text instructions corresponding to each 3D object part, we prepare a set of templates, for example, "Which part of this object <does this function/looks like this>?". We leverage GPT-3.5 \cite{openai2023gpt4} for building instructions using these templates by supplying information about the part and rephrasing. We also extract 3D information from the point cloud, thus designing the instructions to cover relations, dimensions (length, height), comparisons, colour, texture features, object concepts, and functions.  This was further verified manually to avoid inaccurate prompts at test time. The steps are illustrated in Figure \ref{fig:dataanno}.

To cover 3D object segmentation tasks effectively, our text instructions consist of two types: 1) normal queries; 2) 3D queries; as illustrated in Figure \ref{fig:examplesquery}.  
The dataset is partitioned into train and test splits, containing 718 and 1906 3D objects and over 16k and 47k instructions, respectively. As the primary purpose of the dataset is evaluation, the testing set includes a larger number of instructions. 

\subsection{Reasoning 3D Part Segmentation Architecture}
\label{subsec:pipeline}
Our method takes as input a dense and coloured 3D point cloud of an object. One of the common methods of 3D analysis is predominantly using point clouds to represent 3D data. However, this contrasts with human spatial reasoning processes. Humans typically engage with their surroundings through active exploration, synthesizing perspectives from multiple vantages to form an integrated 3D understanding, rather than processing a 3D environment at one go. Our approach advocates for 3D reasoning derived from multi-view imagery. This approach also benefits from the large-scale 2D pretraining available in vision-language models, similar to previous methods that have taken advantage of pre-trained vision-language models for 3D vision tasks. Thus, we render multiple images $x_{img}$ from K predefined camera poses by rasterization. The camera poses cover all parts of the object since they are uniformly distributed around the input point cloud. Given a complex text instruction $x_{txt}$ along with the images, we feed them into a multimodal LLM, denoted by $F$, and the visual backbone, $F_{enc}$. $F$ outputs a text response $\hat{y}_{txt}$. In parallel,  $F_{enc}$ extracts the visual embeddings, f, from each of the input images $x_{img}$. The formulation is as follows: 
\begin{equation}
    \hat{y}_{txt} = F(x_{img}, x_{txt}), 
    f = F_{enc}(x_{img})
\end{equation}
Whenever the LLM is expected to yield a binary segmentation mask, the resultant text output, denoted as  $\hat{y}_{txt}$ is required to an extra token, which sends a request for the segmentation output. Following this, the embedding at the last layer, $\hat{h}_{seg}$, corresponding to the additional segmentation token is extracted. This embedding is subsequently processed through an MLP (Multilayer Perceptron) projection layer, represented as $\gamma$, to derive $h_{seg}$ similar to \cite{lai2023lisa}. The process is as follows: 
\begin{equation}
    h_{seg} = \gamma(\hat{h}_{seg})
\end{equation}
\begin{figure}[t]
    \centering    \includegraphics[width=1\linewidth]{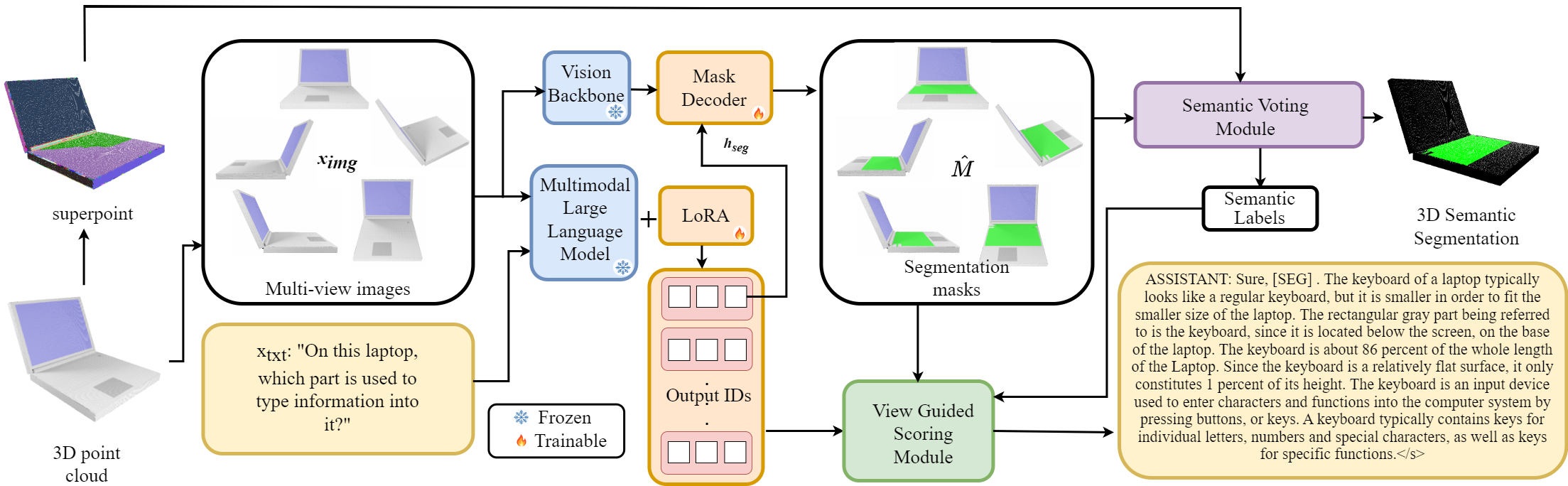}
    \caption{\textbf{Overview of the proposed reasoning-based 3D part segmentation approach named PARIS3D.}  It comprises four subsequent steps: \textbf{(i)}  The 3D point cloud is rendered into K multi-view images $x_{img}$ using a renderer. \textbf{(ii)} These images are passed through a frozen vision backbone (\emph{$F_{enc}$}) and multimodal large language model (\emph{F}) of the reasoning module.  \emph{F} also accepts the text query $x_{txt}$, and produces text outputs corresponding to each view. \textbf{(iii)} The decoder decodes the final layer embedding which contains the extra token, thus producing K segmentation masks.  \textbf{(iv)} Finally, a mask to 3D segmentation algorithm lifts the projections back into 3D and  a view-guided scoring module is used to obtain the final text response. 
     \label{fig:architecure}
  }
\end{figure}
The ensuing step involves the integration of $h_{seg}$ and f into the decoder, $F_{dec}$, which is then responsible for generating the final segmentation mask, $\hat{M}$. The architectural specifics of the decoder, $F_{dec}$, are in accordance with \cite{kirillov2023segment}.

\begin{equation}
    \hat{M} = F_{dec}(h_{seg}, f)
\end{equation}
Once the K segmentations are obtained from each camera angle, a 3D semantic voting module computes scores and assigns the semantic labels for each part. We have $K$ semantic segmentation masks $M_{k}$, where $k$ is the view from which the image was rendered from the point cloud. We aggregate the masks from multiple views and lift them to obtain a three-dimensional semantic segmentation of the original point cloud. To achieve this, we segment the input point cloud P into a set of superpoints {$SP_{i}$}, similar to \cite{landrieu2018largescale}. Superpoints refer to a way of representing large 3D point clouds as a collection of interconnected geometrically simple shapes. This representation is advantageous because (i) it considers complete object parts as a whole, making them easier to identify, (ii) this method can provide a detailed description of the relationship between adjacent objects, and (iii) the size of superpoints is determined by the number of simple structures in a scene, rather than the total number of points, making the representation several orders of magnitude smaller, thus more efficient to work with. Each superpoint contains points with similar normals and colours, suggesting they belong to the same instance. This superpoint-based part labelling not only conserves computational resources but also potentially improves performance by leveraging 3D priors.

In the 3D semantic voting step, we assign semantic labels for each superpoint through a voting mechanism similar to \cite{Liu_2023_CVPR}. This technique leverages information from multiple views and the superpoints, so that even if one mask covers irrelevant points, the aggregation of masks from all views counters the effect of such errors. For a superpoint $SP_{i}$ and a part category j, a score $s_{i,j}$ is computed reflecting the proportion of visible points present in the predicted segmentation masks of category j in each view: 
\begin{equation}
    s_{i,j} = \frac{\sum_{k} \sum_{p \in SP_{i}} [V_{k}(p)][\exists b \in M_{k}^{j} : I_{m}(p)]} {\sum_{k} \sum_{p \in SP_{i}} [V_{k}(p)]}
\end{equation}

Here, [·] is the Iverson bracket, which turns true predicates into 1 and false into 0. $M_{k}^{j}$, represents the predicted mask of category j in view k. $V_{k}(p)$ indicates the visibility of 3D point p in the view k. $I_{m}(p)$ checks if the projection of point p in view k falls within the mask m. The superpoint is then assigned the semantic label of the part category with the highest score.

To provide the most relevant text explanation corresponding to the 3D masks, we propose a view-guided scoring module. It assigns a score to each view based on its correspondence with the final semantic labels. The best text explanation is chosen based on the highest score. Specifically, for each view, we compute the intersection over union (IoU) of the mask with the projections of the final 3D semantic labels output by the model, and the corresponding text explanation is saved. This is called the view-guided score of the explanation. At the end of the score computation for all the masks, we choose the text explanation with the highest score, and in turn, the most correspondence with the output labels. 

\subsection{Training the PARIS3D Architecture}
\subsubsection{Training Data Formulation}
\label{subsec:traindata}
The training data consists of 718 3D objects rendered into multiview images, resulting in over 16k image-instruction pairs. Out of the 718, 360 objects with their instructions are used for training the model and 358 objects with their instructions are used for validation. Each image is provided with an annotation file, which has instructions corresponding to the 3D point cloud that the image came from, the name of the image, and its ground truth mask. Generation of instructions follows the same steps as Section \ref{subsec:dataset} illustrated in Figure \ref{fig:dataanno}. Images in the training set may have more than one instruction in its "instruction" field. This is helpful to introduce diversity as users may randomly select one as the reasoning query during training, thus obtaining a better model.

\subsubsection{Distillation-Based Explanation Refining}
Training the multiview model requires explanatory data as well as training image-instruction pairs. To build the explanatory data, we built an annotation pipeline using a distillation approach. Using the multi-view images as input to the teacher model \cite{lai2023lisa}, we generate explanations for each part of the object. These explanations serve as our pseudolabels which we use as ground truth explanations of the student model. We augment these annotations with 3D features extracted from the 3D point clouds from which the images were rendered. The 3D feature-augmented explanations contain appropriate responses and critical elements differentiating object parts from each other such as location, size, shape, material, and colour. 

\subsubsection{Objective}
\label{sec:training}
We leverage LoRA \cite{hu2022lora} to perform efficient fine-tuning of the pre-trained multimodal LLM $F$ \cite{liu2023llava} to retain its generalization capability. We completely freeze the vision backbone $F_{enc}$. The decoder $F_{dec}$ is fully fine-tuned. The word embeddings of the LLM and the projection layer of $\gamma$ are also trainable, allowing the model to learn the specific meanings and semantic concepts of parts. The model is trained end-to-end using the text generation loss $L_{txt}$ and the segmentation mask loss $L_{mask}$. The overall objective is given as:
\begin{equation}
    L = \lambda_{txt}L_{txt} + \lambda_{mask}L_{mask}
\end{equation}
 Here, $L_{mask}$ encourages the model to produce high-quality segmentation results. It is a combination of per-pixel binary cross-entropy (BCE) loss and DICE loss, with corresponding loss weights $\lambda_{bce}$ and $\lambda_{dice}$, given by: 
 \begin{equation}
    L_{mask} = \lambda_{bce}BCE( \hat{M} ,M) + \lambda_{dice}DICE( \hat{M} ,M)
\end{equation} where $y_{txt}$ and M are the ground-truth targets.  
$L_{txt}$ is the auto-regressive cross-entropy loss for text generation. It is computed as: 
 \begin{equation}
 L_{txt} = CE(\hat{y}_{txt}, y_{txt})
\end{equation}

\section{Experiments}
We perform quantitative and qualitative evaluation of PARIS3D on our dataset for reasoning-based semantic segmentation. 

\subsubsection{Implementation Details and Metrics}
\label{subsec:impdet}
For the experiments, we follow \cite{lai2023lisa} where the multimodal LLM F is LLaVA-13B-v1-1 \cite{liu2023llava} and the vision backbone $F_{enc}$ is the ViT-H SAM. The projection layer of $\gamma$ is an MLP with channels of [256, 4096, 4096]. We use our dataset RPSeg3D which contains coloured point clouds and rendered 2D images of them. Using Pytorch3D \cite{ravi2020accelerating}, each input point cloud is rendered into K = 10 colour images. The fine-tuning scripts for the LLaVA and SAM architecture are based on DeepSpeed \cite{aminabadi2022deepspeed} engine. We adopt the settings of \cite{lai2023lisa} for the optimizer (AdamW \cite{loshchilov2019decoupled}) and its learning rate (0.0003). Similar steps are followed for the learning rate scheduler (WarmupDecayLR), text generation loss $\lambda_{txtgen}$ weight (1.0) and the mask loss $\lambda_{mask}$ weight (1.0), the BCE loss $\lambda_{bce}$ (2.0), dice loss $\lambda_{dice}$ (0.5), batch size per device (2), and the gradient accumulation step (10). The semantic segmentation metric used is category mIoU, following \cite{Liu_2023_CVPR}. It is calculated as follows: first, mIoU for each part category is calculated for all test objects. Then, part mIoUs that belong to each object category are averaged to compute the object category mIoU.

\subsection{Reasoning Part Segmentation Results}
\label{subsec:reasonpartresult}
We establish the reasoning part segmentation task on our dataset RPSeg3D. Table \ref{tab:reason-seg} shows the results of reasoning part segmentation. We observe that without any fine-tuning, the model's performance is low compared to its fine-tuned counterparts. The general observation is that 3D queries help the model to output better masks. We compare our method to two baselines. The first baseline is LISA \cite{lai2023lisa} applied to the multiple views without any finetuning. The second baseline consists of LISA finetuned on few-shot part segmentation data. When fine-tuned with 3D information, our model performs better than the baselines for normal and 3D prompts. 
\begin{table}[t]
\centering
\resizebox{.8\textwidth}{!}{%
\begin{tabular}{l|lll|lll}
\hline
\multirow{2}{*}{Method} & \multicolumn{3}{l|}{Val}                         & \multicolumn{3}{c}{Test}                               \\ \cline{2-7} 
 & \multicolumn{1}{l|}{Normal Query} & \multicolumn{1}{l|}{3D Query} & Overall & \multicolumn{1}{l|}{Normal Query} & \multicolumn{1}{l|}{3D Query} & Overall \\ \hline
LISA-MV           & \multicolumn{1}{l|}{16.60}  & \multicolumn{1}{l|}{20.16} &  18.38 & \multicolumn{1}{l|}{17.60} & \multicolumn{1}{l|}{20.57} & 19.08 \\
LISA-MV (ft)       & \multicolumn{1}{l|}{50.43} & \multicolumn{1}{l|}{50.28} & 50.35 & \multicolumn{1}{l|}{50.75} & \multicolumn{1}{l|}{50.81} & 50.78 \\
PARIS3D         & \multicolumn{1}{l|}{55.33} & \multicolumn{1}{l|}{55.50} & 55.42 & \multicolumn{1}{l|}{55.94} & \multicolumn{1}{l|}{57.60} & 56.77 \\ \hline
\end{tabular}%
}
\caption{\textbf{Results of reasoning part segmentation.} LISA-MV \cite{lai2023lisa} is LISA in multi-view setting without fine-tuning. LISA-MV (ft) is the experiment in which it has been fine-tuned on few-shot part segmentation data. Our proposed PARIS3D method has been fine-tuned with 3D queries and explanations. When fine-tuned with 3D information, our model performs better than the baselines for normal and 3D prompts. Here \textit{Test} is the test set of 1906 shapes and \textit{Val} is the validation set of 358 3D shapes with their instructions. }
\label{tab:reason-seg}
\end{table}

\subsection{3D Semantic Segmentation Comparison with Existing Models}
\label{subsec:3dsemseg}
Table \ref{tab:sem-seg} shows the results of semantic segmentation compared to existing methods. Our method has better performance than all the fully supervised baselines and achieves competitive results with \cite{Liu_2023_CVPR}. For this baseline, we observe that the model with prompt tuning done separately on each category achieves impressive few-shot performance but when unified into a single model for all categories, the performance is only slightly better than its zero-shot performance (27.2\%). This is attributed to the redundant part names across object categories, hindering the model's learning of the semantic meanings of part names. This results in performance to drop significantly from when only one category is learnt per model. %
To prompt for segmentation, the baseline models \cite{qian2022pointnext, qi2017pointnet, zhao2021few} are provided with the class IDs of the parts to be segmented (e.g: 0 to 103 including the background). In \cite{Liu_2023_CVPR}, the model is provided with the names of the parts to be segmented (e.g. "seat", "arm", "back" of a chair). Our method, PARIS3D, uses language instructions to prompt segmentation. For a fair comparison, we use hand-crafted prompts with short instructions containing generic concept and location-based clues about the part to be segmented. 
\begin{table}[t]
\centering
\resizebox{\textwidth}{!}{%
\begin{tabular}{l|l|lllllllll|lllllllll|l}
\hline
\multirow{2}{*}{3D Data} &
  \multirow{2}{*}{Method} &
  \multicolumn{9}{l|}{Overlapping Categories} &
  \multicolumn{9}{l|}{Non-overlapping Categories} &
  \multirow{2}{*}{\begin{tabular}[c]{@{}l@{}}Overall \\ (45)\end{tabular}} \\ \cline{3-20}
 &
   &
  Bottle &
  Chair &
  Display &
  Door &
  Knife &
  Lamp &
  \begin{tabular}[c]{@{}l@{}}Storage-\\ Furniture\end{tabular} &
  \multicolumn{1}{l|}{Table} &
  \begin{tabular}[c]{@{}l@{}}Overall \\ (17)\end{tabular} &
  Camera &
  Cart &
  \begin{tabular}[c]{@{}l@{}}Dis-\\ Penser\end{tabular} &
  Kettle &
  \begin{tabular}[c]{@{}l@{}}Kitchen\\ Pot\end{tabular} &
  Oven &
  \begin{tabular}[c]{@{}l@{}}Suit-\\ case\end{tabular} &
  \multicolumn{1}{l|}{Toaster} &
  \begin{tabular}[c]{@{}l@{}}Overall \\ (28)\end{tabular} &
   \\ \hline
\multirow{3}{*}{\begin{tabular}[c]{@{}l@{}}Extra data \\ (45x8+28k)\end{tabular}} &
  PointNet++ \cite{qi2017pointnet} & 48.8
   & 84.7
   & 78.4 
   & 45.7
   & 35.4
   & 68.0
   & 46.9
   & 
  \multicolumn{1}{l|}{\textbf{63.7}} & 55.6
   & 6.5
   & 6.4
   & 12.1
   & 20.9
   & 15.8
   & 34.3
   & 40.6
   & 
  \multicolumn{1}{c|}{14.7} & 25.4
   & 36.8
   \\
 &
  PointNeXt \cite{qian2022pointnext} & 68.4 
   & \textbf{91.8} 
   & \textbf{89.4} 
   & 43.8 
   & 58.7 
   & 64.9 
   & \textbf{68.5} 
   & 
  \multicolumn{1}{c|}{52.1} & \textbf{58.5}
   & 33.2 
   & 36.3 
   & 26.0 
   & 45.1 
   & 57.0 
   & 37.8 
   & 13.5 
   &  
  \multicolumn{1}{c|}{8.3} & 45.1
   & 50.2
   \\
 &
  SoftGroup \cite{vu2022softgroup} & 41.4 
   & 88.3 
   & 62.1 
   & 53.1 
   & 31.3 
   & \textbf{82.2} 
   & 60.2 
   & 
  \multicolumn{1}{c|}{54.8} & 50.2 
   & 23.6 
   & 23.9 
   & 18.9 
   & 57.4 
   & 45.5 
   & 13.6 
   & 18.3 
   &  
  \multicolumn{1}{c|}{26.4} & 30.7 
   & 38.1
   \\ \hline
\multirow{8}{*}{\begin{tabular}[c]{@{}l@{}}Few-shot \\ (45x8)\end{tabular}} 
 &
  PartSLIP* \cite{Liu_2023_CVPR} & 83.4 
   & 85.3 
   & 84.8 
   & 40.8 
   & \textbf{65.2}
   & 66.0 
   & 53.6 
   & 
  \multicolumn{1}{c|}{42.4} & 56.3
   & \textbf{58.3}
   & \textbf{88.1}
   & \textbf{73.7} 
   & \textbf{77.0} 
   & 69.6 
   & \textbf{73.5} 
   & \textbf{70.4}
   &  
  \multicolumn{1}{c|}{\textbf{60.0}} & \textbf{61.3} 
   & \textbf{59.4}
   \\
&
  PointNet++ \cite{qi2017pointnet} & 27.0 
   & 42.2 
   & 30.2 
   & 20.5 
   & 22.2 
   & 10.5 
   & 8.4 
   & 
  \multicolumn{1}{c|}{7.3} & 18.1 
   & 9.7 
   & 11.6 
   & 7.0 
   & 28.6 
   & 31.7 
   & 19.4 
   & 3.3 
   &  
  \multicolumn{1}{c|}{0.0} & 21.8 
   & 20.4
   \\ 
 &
  PointNeXt \cite{qian2022pointnext} & 67.6 
   & 65.1 
   & 53.7 
   & 46.3 
   & 59.7 
   & 55.4 
   & 20.6 
   & 
  \multicolumn{1}{c|}{22.1 } & 39.2 
   & 26.0 
   & 47.7 
   & 22.6 
   & 60.5 
   & 66.0 
   & 36.8 
   & 14.5 
   & 
  \multicolumn{1}{c|}{0.0} & 41.5 
   & 40.6
   \\
 &
  SoftGroup \cite{vu2022softgroup} & 20.8 
   & 80.5 
   & 39.7 
   & 16.3 
   & 38.3 
   & 38.3 
   & 18.9 
   & 
  \multicolumn{1}{c|}{24.9 } & 32.8 
   & 28.6 
   & 40.8 
   & 42.9 
   & 60.7 
   & 54.8 
   & 35.6 
   & 29.8 
   & 
  \multicolumn{1}{c|}{14.8 } & 41.1 
   & 38.0
   \\
 &
  ACD \cite{selfsupacd} & 22.4 
   & 39.0 
   & 29.2 
   & 18.9 
   & 39.6 
   & 13.7 
   & 7.6 
   & 
  \multicolumn{1}{c|}{13.5} & 19.2 
   & 10.1 
   & 31.5 
   & 19.4 
   & 40.2 
   & 51.8 
   & 8.9 
   & 13.2 
   &  
  \multicolumn{1}{c|}{0.0} & 25.6 
   & 23.2
   \\
 &
  Prototype \cite{zhao2021few} & 60.1 
   & 70.8 
   & 67.3 
   & 33.4 
   & 50.4 
   & 38.2 
   & 30.2 
   & 
  \multicolumn{1}{c|}{25.7} & 41.1 
   & 32.0 
   & 36.8 
   & 53.4 
   & 62.7 
   & 63.3 
   & 36.5 
   & 35.5 
   &  
  \multicolumn{1}{c|}{10.1} & 46.3 
   & 44.3
   \\

 &
  PartSLIP+ & 64.8
   & 69.5
   & 59.5
   & 24.5
   & 34.5 
   & 37.1
   & 32.0
   & 
  \multicolumn{1}{c|}{40.1} & 35.3
   & 25.5
   & 75.7
   & 15.6
   & 30.5
   & 58.4
   & 31.1
   & 49.4
   & 
  \multicolumn{1}{c|}{6.6} & 26.7
   & 29.9
   \\
   \rowcolor{Lavender}
 &
 
  Ours & \textbf{84.0}
   &  81.0
   &  70.1
   & \textbf{68.4}
   & 47.2
   & 61.2
   & 39.4
   & 
  \multicolumn{1}{c|}{45.1} & 55.1
   & 29.3
   & 71.7
   & 40.1
   & 59.3
   & \textbf{78.8}
   & 59.1
   & 61.6
   & 
  \multicolumn{1}{c|}{24.9} & 59.1
   & 57.6
   \\ \hline
   
\end{tabular}%
}
\caption{Comparison to previous 3D part segmentation methods. Object category mIoU(\%) is shown. In the 45x8+28k setting, baseline models use an additional 28k training shapes for 17 overlapping object categories. These are categories present in common with PartNet datset. For the remaining 28 non-overlapping object categories, there are only 8 shapes per object category during training. PartSLIP* indicates that \textbf{one model has been trained for each category}. + shows our implementation of PartSLIP where one model is trained for all the categories together.}
\label{tab:sem-seg}
\end{table}
\begin{figure}[t]
\centering
    \includegraphics[width=1\linewidth]{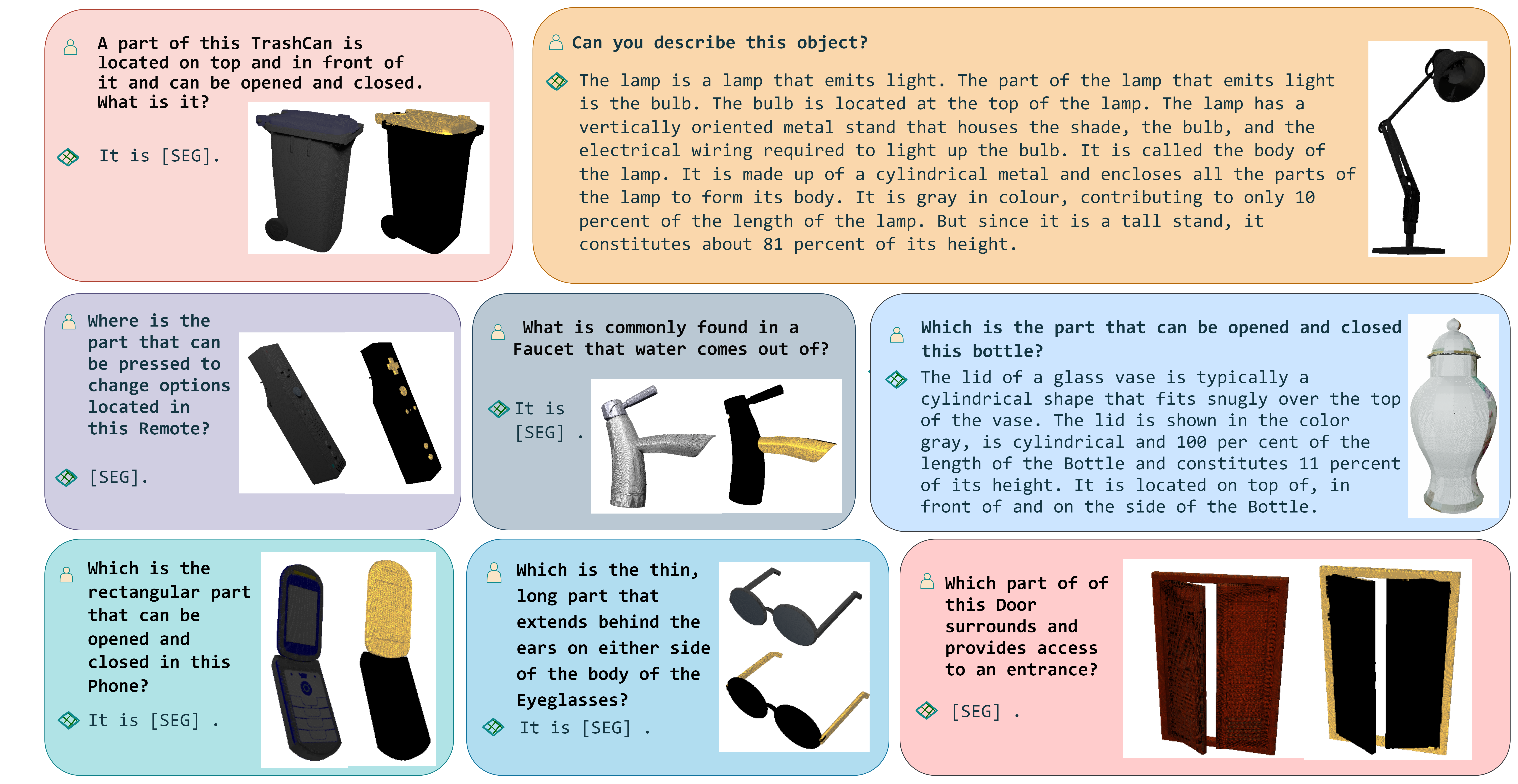}
    \caption{\textbf{Qualitative results of PARIS3D's performance}. We showcase examples from three tasks: reasoning 3D object part segmentation, object description, and reasoning question-answering, demonstrating its capabilities in offering in-depth reasoning, 3D understanding, part segmentation, and conversational abilities.}
    \label{fig:qualitative}
\end{figure}
\subsection{Ablation Study}
\noindent \textbf{Instruction Rephrasing.}
 \label{sec:3dquery}
  We use GPT-3.5 \cite{openai2023gpt4} to generate reasoning instructions corresponding to each part of 45 object categories for training the model. We further enrich them with colour information. As opposed to fine-tuning on purely reasoning data and/or colours, fine-tuning on 3D-related information such as position, shape and dimensions was more useful for semantic segmentation as observed in Table \ref{tab:3dquery}. 
\begin{table}[]
\centering
\resizebox{.9\textwidth}{!}{
\centering
\begin{tabular}{c|cccccc}
\hline
Instructions   & KitchenPot & Phone & Keyboard & Oven & StorageFurniture & Overall \\ 
          & lid        & lid     & cord   & door & door & \\ 
\hline
GPT-generated & 45.8  & 24.8  &  56.3 & 57.2  & 39.8   & 46.7\\
GPT + Colour  & 79.2  & 38.9  & 96.6  & 65.2  & 42.4 & 50.8\\ 
GPT + Colour + 3D information  & \textbf{85.9} & \textbf{61.6} & \textbf{99.0} & \textbf{74.3} & \textbf{45.3} & \textbf{57.6}\\
\hline
\end{tabular}
}
\caption{Results of training the model on different incremental prompts. \textit{GPT-generated} means that the training data consists of multiple prompts rephrased by GPT-3.5 related to the part concept or function. In the second setting, \textit{GPT + Colour}, these prompts are infused with colour information about the 3D train shapes. The final experiment has the model trained on multiple rephrased prompts, colour and other 3D information extracted from the training point clouds. }
\label{tab:3dquery}
\hfill
\end{table}
 
\noindent \textbf{One Model vs Multiple Models.}
\label{sec:45mod}
One of the baseline models \cite{Liu_2023_CVPR} trained a model for each of the 45 categories, loading one at a time to evaluate each category. When we replicate this setting for PARIS3D, there are significant jumps in performance as shown in Table \ref{tab:45model}. However, one-model-for-one-category does not offer a generalizable solution in a real-world problem setting, where multiple object categories and their parts need to be analysed.
\begin{table}[t]
\centering
\begin{tabular}{c|ccccc}
\hline
Method   & Chair & Dispenser & Keyboard & Eyeglasses & Bucket \\ \hline
PartSLIP \cite{Liu_2023_CVPR} & 85.3  & 73.7      & 53.6     & 88.3       & 36.5   \\
PARIS3D  & \textbf{86.5} & \textbf{75.6} & \textbf{88.2} & \textbf{92.2} & \textbf{83.9}\\ 
\hline
\end{tabular}
\caption{Results of PARIS3D on one model trained for each of 5 categories. In \cite{Liu_2023_CVPR}, 45 models were trained and each point cloud was tested by loading its corresponding model to perform evaluation. Repeating this exact setting by training a model on each category for PARIS3D, we easily gain +1.2\%, +1.9\%, +34.6\%, +3.9\%, and +47.4\% improvements on the tested categories.}
\label{tab:45model}
\hfill
\end{table}
\\
\noindent \textbf{Number of Rendered Views.}
The effect of incrementing the number of views rendered from the input point cloud is shown in Table \ref{tab:views}.
With only 1 view information provided, the segmentations are not meaningful. With 5 views, the segmentation quality improves, and improves further with an increase in the number of views. For each point cloud, we use 10 rendered views in our main experiments. \\
\noindent \textbf{Single prompt vs Multiple prompts.}
\label{sec:multiprompt}
In Table \ref{tab:multiprompt}, we show ablation experiments regarding the number of training prompts. Single prompt refers to one concept-based reasoning training instruction that follows a template. In further experiments, we use the template to generate multiple (about 5-6) rephrased instructions for each part category using \cite{openai2023gpt4}. Training with these multiple training prompts per part category results in a better-performing model than with a single prompt. 
\begin{table}
\parbox{.45\linewidth}{
\centering
\begin{tabular}{cccc}
\hline
No. of views & Bottle & Bucket & Camera\\
\hline
1 & 1.1 & 1.1 & 1.2\\
5 & 23.7 & 40.1 & 12.1\\
10 & 84.8 & 84.9 & 29.3\\
\hline
\end{tabular}
\caption{Ablation experiments on the number of rendered views. }
\label{tab:views}
}
\parbox{.45\linewidth}{
\centering
\begin{tabular}{ccccc}
\hline
No. of prompts & Bottle & Knife & Clock & Chair\\
\hline
1 & 73.8 & 44.3 & 45.5 & 69.5\\
5-6 & \textbf{83.9} & \textbf{52.7} & \textbf{51.3} & \textbf{78.7}\\
\hline
\end{tabular}
\caption{Ablation experiments on the number of prompts provided in training data. }
\label{tab:multiprompt}
}

\end{table}
\subsection{Qualitative Results}
In Figure \ref{fig:qualitative}, we provide examples of PARIS3D output for 3D object part segmentation tasks. PARIS3D is capable of segmenting fine-grained details in categories like buttons of a remote or legs of an eyeglass from implicit queries that expect it to do self-reasoning to generate the right answer, and then segment the part. Moreover, it can perform tasks like object description and question-answering with regard to object parts and give detailed answers. This demonstrates its conversational, reasoning, and part segmentation capabilities.

\subsection{Generalizability to Real-World Examples}
\begin{figure}[t]
\centering
\includegraphics[width=1\linewidth]{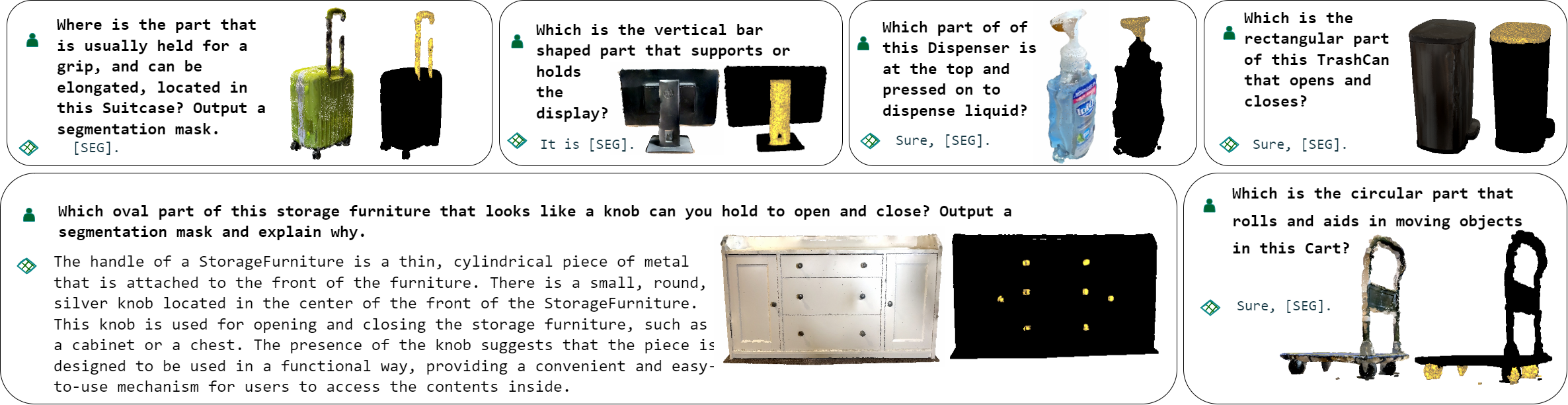}
\caption{\textbf{Qualitative results on real-world examples of point clouds.} The input point cloud is shown on the left and the segmentation mask is in yellow on the right. Our model generalizes to point clouds with a significant domain gap from those it was trained on and displays impressive results with segmenting small object parts such as the furniture's knob in this example. }
    \label{fig:realpc}
\end{figure}

Most existing 3D tasks and their corresponding models can usually only handle cases within the same distribution of the training sets without generalization, since they are sensitive to the format of the input and significant domain gap between synthetic experiments and real-world examples. Thus, it can be difficult to use them in use cases involving real point clouds from an everyday setting. To demonstrate the generalizability of PARIS3D to data derived from the real world, we perform our 3D segmentation on real point clouds shot using a smartphone's LiDAR sensor, as suggested by \cite{Liu_2023_CVPR}. In Figure \ref{fig:realpc}, we show qualitative examples of passing the fused point clouds through the PARIS3D architecture to obtain part segmentation labels as in the previous experiments without much drop in performance.
\section{Conclusion}
In this work, we introduce a novel challenge within 3D segmentation, reasoning-based part segmentation. This task requires models to infer, reason, and explain based on implicit user instructions, making it considerably more complex than the regular 3D referring segmentation task. We introduce a dataset for this task, RPSeg3D, to enable effective evaluation. We believe this dataset will play a crucial role in fostering the growth of technologies in this area. Additionally, we outline a pipeline that integrates 3D segmentation capabilities into multimodal Large Language Models (LLMs), showcasing our model, PARIS3D, which exhibits competitive performance. It additionally demonstrates the ability to identify part concepts, reason about them, and complement them with world knowledge. However, we identify limitations - the model in its current form cannot perform instance segmentation. This is a direction for future research as we expand the dataset to accommodate such tasks. 

\clearpage
\bibliographystyle{splncs04}
\bibliography{egbib}
\end{document}